\documentclass{article}

\usepackage{microtype}
\usepackage{graphicx}
\usepackage{subfigure}
\usepackage{booktabs} 

\usepackage{hyperref}

\usepackage[accepted]{icml2024}

\usepackage{amsmath}
\usepackage{amssymb}
\usepackage{mathtools}
\usepackage{amsthm}

\usepackage[capitalize,noabbrev]{cleveref}

\theoremstyle{plain}

\theoremstyle{definition}

\theoremstyle{remark}

\usepackage[textsize=tiny]{todonotes}

\usepackage[utf8]{inputenc} 
\usepackage[T1]{fontenc}    
\usepackage{hyperref}       
\usepackage{url}            
\usepackage{booktabs}       
\usepackage{amsfonts}       
\usepackage{nicefrac}       
\usepackage{subfigure}
\usepackage{microtype}      
\usepackage{multirow}
\usepackage{xcolor}         
\usepackage{algorithm}
\usepackage{algorithmic}
\usepackage{mdframed}
\usepackage{amsmath}
\usepackage{listings}
\usepackage{amssymb}
\usepackage{graphicx}
\usepackage{wrapfig}

\definecolor{MyDarkBlue}{rgb}{0,0.08,1}
\definecolor{MyDarkGreen}{rgb}{0.015,0.45,0.015}
\definecolor{MyDarkRed}{rgb}{0.8,0.02,0.02}
\definecolor{MyDarkOrange}{rgb}{0.80,0.4,0.04}
\definecolor{MyPurple}{RGB}{111,0,255}
\definecolor{MyRed}{rgb}{1.0,0.0,0.0}
\definecolor{MyGold}{rgb}{0.75,0.6,0.12}
\definecolor{MyDarkgray}{rgb}{0.66, 0.66, 0.66}

\renewcommand{\paragraph}[1]{\noindent \textbf{#1}~}

\newcommand{\dual}[1]{\multirow{2}{*}{#1}}
\newcommand{\ul}[1]{\underline{#1}}

\newcommand{\sidistoryline}[1]{}

\icmltitlerunning{Open-Domain Text Evaluation via Contrastive Distribution Methods}

\begin{document}
\twocolumn[
\icmltitle{Open-Domain Text Evaluation via Contrastive Distribution Methods}
\icmlsetsymbol{equal}{*}

\begin{icmlauthorlist}
\icmlauthor{Sidi Lu}{ucla}
\icmlauthor{Hongyi Liu}{sjtu}
\icmlauthor{Asli Celikyilmaz}{meta}
\icmlauthor{Tianlu Wang}{meta}
\icmlauthor{Nanyun Peng}{ucla}
\end{icmlauthorlist}

\icmlaffiliation{ucla}{Department of Computer Science, University of California, Los Angeles}
\icmlaffiliation{sjtu}{Shanghai Jiao Tong University}
\icmlaffiliation{meta}{Meta FAIR}

\icmlcorrespondingauthor{Sidi Lu}{sidilu@cs.ucla.edu}
\icmlcorrespondingauthor{Nanyun Peng}{violetpeng@cs.ucla.edu}

\icmlkeywords{Language Models, Text Evaluation, ICML}

\vskip 0.3in
]
\printAffiliationsAndNotice{}
\begin{abstract}
Recent advancements in open-domain text generation, driven by the power of large pre-trained language models (LLMs), have demonstrated remarkable performance. However, assessing these models' generation quality remains a challenge. 
In this paper, we introduce a novel method for evaluating open-domain text generation called Contrastive Distribution Methods (CDM). Leveraging the connection between increasing model parameters and enhanced LLM performance, CDM creates a mapping from the \textit{contrast} of two probabilistic distributions -- one known to be superior to the other -- to quality measures. We investigate CDM for open-domain text generation evaluation under two paradigms: 1) \emph{Generative} CDM, which harnesses the contrast of two language models' distributions to generate synthetic examples for training discriminator-based metrics; 2) \emph{Discriminative} CDM, which directly uses distribution disparities between two language models for evaluation. Our experiments on coherence evaluation for multi-turn dialogue and commonsense evaluation for controllable generation demonstrate CDM's superior correlate with human judgment than existing automatic evaluation metrics, highlighting the strong performance and generalizability of our approach. 
\footnote{Code: \url{https://github.com/PlusLabNLP/CDM}}

\end{abstract}
 
\section{Introduction}
\sidistoryline{Mention the recent successes of large language models (LLM) like GPT-3/ChatGPT. Briefly explain why proper evaluation of such models is becoming an emerging problem, and talk about the limitations of existing metrics (including statistical metrics like BLEU and pretrained-model-based metrics like BERTScore) in the follow perspectives: 1) metrics like BLEU does not work for diverse distributions 2) metrics like BERTScore fall short for multi-turn or long contents.} 

In recent years, open-domain text generation, fueled by large pretrained generative language models (LLMs), has made significant advancements, garnering substantial attention \citep{radford2018gpt1,radford2019gpt2,brown2020gpt3,openai2022chatgpt,openai2023gpt4}. These systems have showcased remarkable capabilities, such as producing human-like responses, contributing to natural language comprehension, and even performing complex tasks like programming and content generation. With the empirical success, the development of reliable and scalable automatic evaluation metrics for these models become imperative, yet the problem remains an unresolved challenge.


Existing automatic evaluate metrics from pre-LLM eras have their respective limitations. Specifically, reference-based statistical metrics (e.g. BLEU \citep{papineni2002bleu}, ROUGE \citep{lin2004rouge}, METEOR~\citep{banerjee2005meteor}) do not work well for open-ended generation problems with high content diversity like storytelling~\citep{yao2019plan} and dialogue systems \citep{mesgar2019dialogue,li2017adversarial,wen2016network}, as for these tasks, it is challenging, if not impossible, to collect a sufficiently large number of reference examples to represent the distribution of all feasible outputs. Therefore, prior works have shown their low correlation with human judgments~\citep{liu2016not,hu2020makes}.
With recent progress in pretrained models, model-based reference metrics like BERTScore~\citep{zhang2019bertscore}, Bluert~\citep{sellam2020bleurt} are proposed to facilitate automatic evaluation for text generation. They alleviate the sample efficiency issue of statistical reference-based methods by using pretrained models to compute the similarities between texts based on higher-level semantics. However, the effectiveness of such methods is still reliant on the representativeness of the reference set, and thus falls short when the output semantic space is also highly diverse. 

\begin{figure}
  \centering  
    \includegraphics[width=\columnwidth]{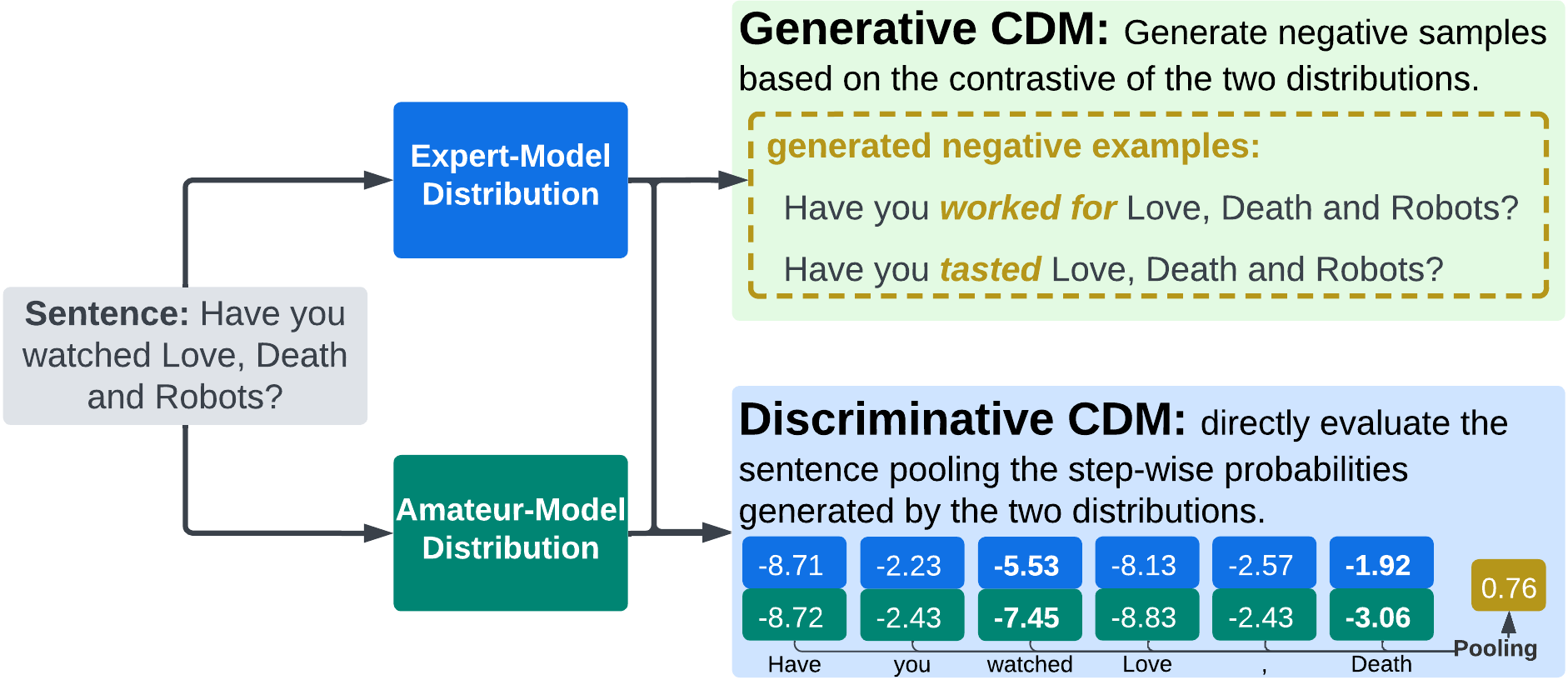} \label{fig:overview-cdm}
  \vspace{-2em}
  \caption{Conceptual illustration of the Contrastive Distribution Methods (CDM). 
  (a) \emph{Generative CDM} generates negative examples for training a discriminator-based metric. (b) \emph{Discriminative CDM} directly evaluate the distribution/sequence by contrasting the step-wise likelihood scores. 
  }
  \vspace{-1.5em} 
\end{figure}

Reference-free evaluation metrics, which assess text directly and provide a quality score, offer a more flexible solution for automatically evaluating open-domain text generation. There are two major paradigms for training models to evaluate texts without references: 1) \emph{Discriminator-based approaches} like ADEM~\citep{lowe2017autoturing} and DEAM~\citep{ghazarian2022deam}, treat the problem as a prediction task. They train a classifier or regressor to generate a score as the quality assess. However, these methods typically require extensive human annotations or involve dedicated manual designs for generating negative samples to train the classifier. 2) \emph{Distribution/Divergence-based approaches} \citep{pillutla2021mauve,pimentel2022usefulness} focus on obtaining a continuous divergence score between distributions. These approaches have shown promising results in system-level evaluations. However, they often face challenges to accurately assigning credit to individual data points, limiting their ability to perform instance-level evaluations. 
In this paper, we propose Contrastive Distribution Methods (CDM), a general and reference-free framework 
for evaluating open-domain text generation. CDM operates on an intuitive yet broadly applicable premise: models with similar architectures but varying sizes generally exhibit improved performance as model size increases. 
Consequently, CDM is designed to 
capture the dynamics of model performance as it scales with the increasing number of parameters. Utilizing such dynamics, CDM \textit{contrasts} two language models' distributions and conduct inference in both \textit{generative} and \textit{discriminative} manners to create automatic evaluation metrics. 
Specifically, Generative CDM as illustrated in the upper right corner of Figure~\ref{fig:overview-cdm} 
produces effective negative samples to facilitate the learning of discriminator-based evaluation metrics without the requirement of additional human annotations or sophisticated design for the data generation process, and Discriminative CDM as illustrated in the lower right corner of Figure~\ref{fig:overview-cdm} 
provides a distribution-level measurement of quality for each instance, and thus results in reliable distribution-based metrics without compromising instance-level evaluation performance. 

Experiments on open-domain dialogue evaluation and commonsense keywords-to-text evaluation demonstrate strong performance of CDMs, consistently outperforming strong baselines such as G-Eval~\cite{liu2023geval} in terms of correlation with human judgements across datasets. \looseness=-1

\section{Background and Related Works} 


\paragraph{Open-Domain Text Evaluation} There has been a synchronously growing interest in developing robust evaluation methods for open-domain text generation models. Traditional evaluation metrics, such as BLEU and ROUGE, have been shown to be inadequate for assessing the quality of complex, multi-sentence responses generated by these models. As a result, researchers have explored alternative evaluation methods, including human evaluation, adversarial evaluation, and unsupervised metrics. Human evaluation remains the gold standard, but it is time-consuming and costly. Adversarial evaluation, which involves testing models against a set of challenging examples, has shown promise in identifying weaknesses in current models. Unsupervised metrics, such as BERTScore and Perplexity, provide quick and automated evaluation, but their correlation with human judgments remains a topic of debate. The field of open-domain text evaluation continues to evolve, and developing reliable evaluation methods will be essential for advancing the state-of-the-art in this exciting area of research.

\paragraph{Discriminator-based Metrics} ADEM \citep{lowe2017autoturing} is one of the first attempts at training a model to evaluate machine-generated text. It deals with single-turn dialogue evaluation problem, and uses the contextualized representation of the context in interaction with that of the responses to train the model. DEAM \citep{ghazarian2022deam} and AMRFact \citep{qiu2023amrfact} are novel evaluation metrics that aim to assess open-end generation models with structured manipulations to create negative samples from positive ones, allowing for a more nuanced assessment of model performance like coherence (for dialogue system) or factuality (for summarization models). Typically, they operate by first parsing the sequence into an abstract meaning representation (AMR), and then manipulating the AMR to introduce inconsistencies and irrelevancies that undermine the coherence of the dialogue. The manipulated AMR is then transformed back into text form for evaluation. This method supports multi-turn dialogue evaluation and has achieved state-of-the-art performance on various benchmark datasets. By using AMR-based semantic manipulations, these methods provide a class of promising approaches for performing automatic evaluation in a more comprehensive and accurate manner. \emph{Generative} CDM shares a similar process
, as it manipulates the positive true samples for the generation of negative samples, serving the purpose of training a classifier.

\paragraph{Distribution/Divergence-based Metrics} MAUVE and follow-up works \citep{pillutla2021mauve,pimentel2022usefulness} analyse the quality gap between human-generated text and machine-generated text by studying the divergence frontier of human-generated samples in contrast to the learnt model. While their setup is not directly relevant to our approach, it provides an insightful perspective of using the likelihood predictions of LMs for evaluation purposes. \citet{zhong2022towards} proposes a multi-dimensional evaluation system for more robust automatic evaluation. It ensembles the score from a set of \emph{discriminator-based} metrics, each of which trained to evaluate a specific aspect in intuition of the text quality. GPTEval
\citep{liu2023gpteval} tries to quantitatively exploit large language models that are trained with strong human alignment. It uses the score prediction from GPT-4 \citep{openai2023gpt4} to evaluate how well the given text adheres to human opinion. \emph{Discriminative} CDM falls under this paradigm, since it serves as a metric with more continuously distributed scores for the evaluated text.

\paragraph{Contrastive Decoding, Contrastive Momentum and ExPO} Contrastive decoding is a decoding algorithm that leverages the strengths of two language models: a stronger expert model and a weaker amateur model. The algorithm decodes towards the objective of maximizing the difference between the log-probabilities of the expert and amateur models, resulting in high-quality generated samples. Specifically, the algorithm tries to decode sequences that maximize the \emph{contrastive momentum}:
\begin{equation}
    \log p_{\text{e}}(x) - \log p_{\text{a}}(x), 
\end{equation}
where $p_{\text{e}}$ and $p_{\text{a}}$ represent the expert and the amateur models, respectively, and $x$ is the generated sample. The original paper \citep{li2022contrastive} 
demonstrates that this approach results in higher quality samples than decoding from the expert model alone. Contrastive decoding provides an insightful way to study the dynamics of how models' capabilities scale up with larger parameter numbers. The proposed CDM is highly inspired by the Contrastive decoding method, yet leveraging it for evaluation purposes.

Noticeably, a recent preference optimization method called ExPO~\citep{zheng2024weak} also shares a similar idea. ExPO significantly improves the instruction following abilities of (open-sourced) large language models without the necessity of performing any costly training and with even less data and/or trial sampling from the LLMs. It creates the extrapolation of human-aligned models (in our notion, the \emph{expert}) in contrast to its primitive version after only the supervised finetuning (SFT) stage (in our notion, the \emph{amateur}) on instruction-following data. The biggest difference between ExPO and contrastive decoding or this paper is, since the \emph{amateur} and \emph{expert} models in ExPO share the same parameter space and is assumed to be very close to each other, ExPO directly performs the extrapolation in the parameter space, instead of the log-probability space as in ours or the contrastive decoding algorithm.

\section{Methodology}

\subsection{Notations and Problem Formulation}
\label{sec:notations}
We use $\mathbf{s}$ to denote a sequence and $s_i$ to denote the $i-$th token in $\mathbf{s}$. $p(\mathbf{s})$ denotes the probability of sequence $\mathbf{s}$ under a model $p$. 
We assume model $p$ is a probabilistic distribution defined on $\Sigma^*$, where $\Sigma$ is the set of valid tokens and $\Sigma^*$ is the universal set of all sequences consisting of such tokens. 
        
Consider an imaginary distribution-level oracle metric $E(p)$ 
which projects from a model distribution $p(\mathbf{s})$ 
to ``a measure of model performance'' -- a scalar. This function does not necessarily have an analytical form, 
however, we assume that we have access to some partial order relations it defines. 
Intuitively, this imaginary oracle $E(p)$ should correlate perfectly with human judgements of the model performance, and any evaluation metric that correlates better with human judgments is a better approximation of $E(p)$. 
\begin{figure}[t] 
  \centering
  \subfigure[]{\includegraphics[height=0.6\columnwidth]{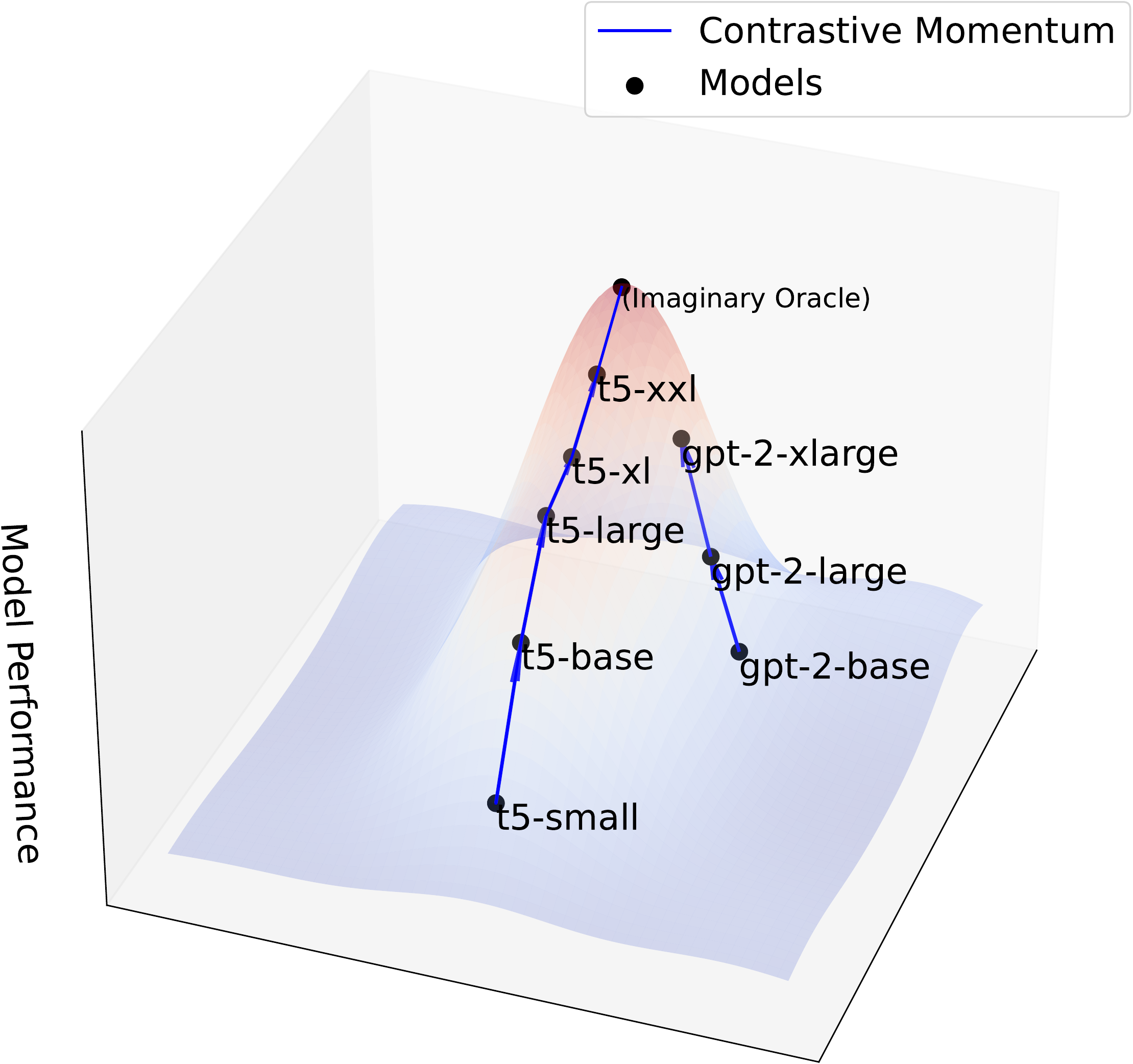} \label{fig:secant}}
  
  \centering
    \subfigure[]{\includegraphics[height=0.3\columnwidth]{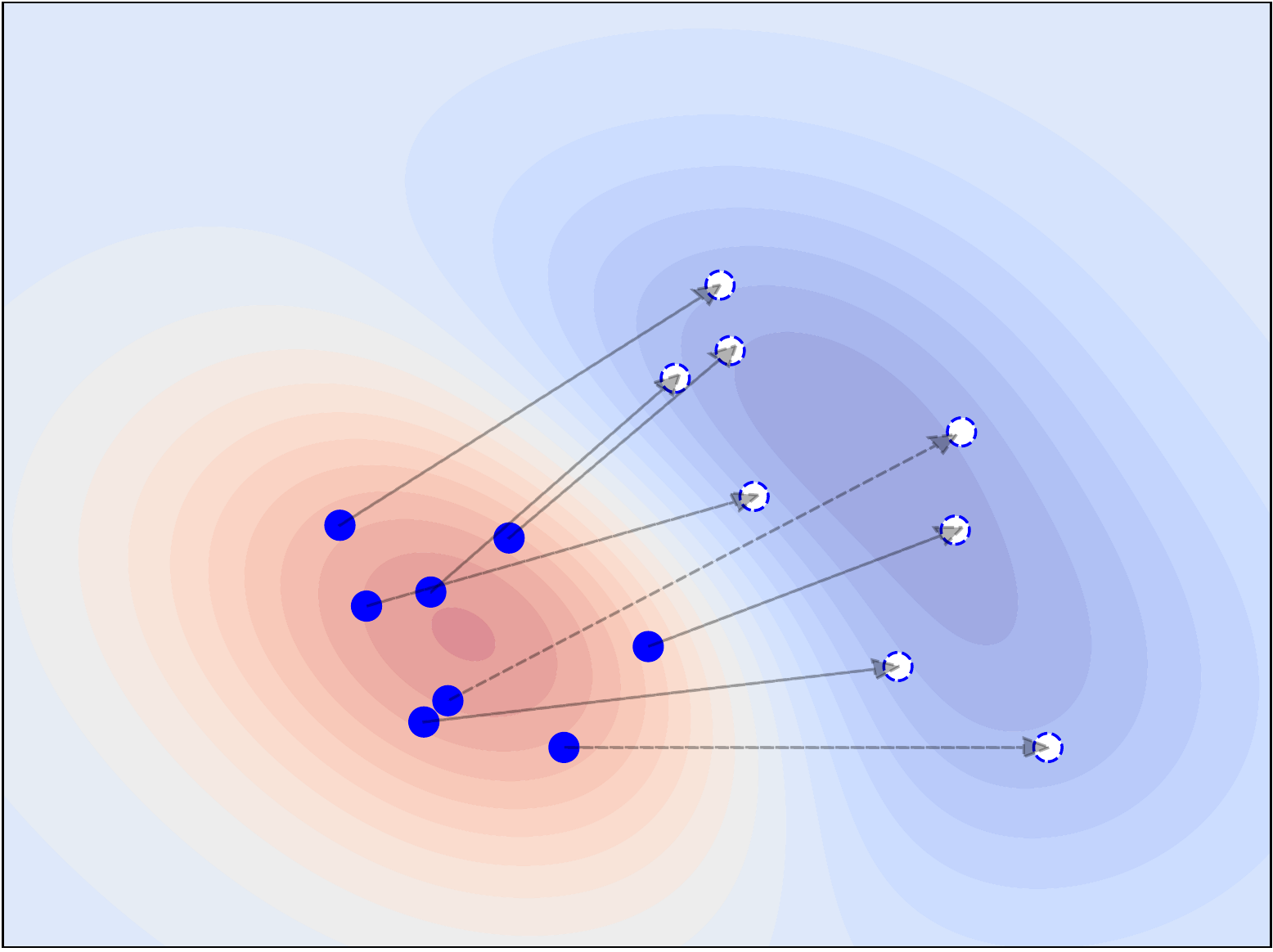}\label{fig:generative_CDM}}
    \subfigure[]{\includegraphics[height=0.3\columnwidth]{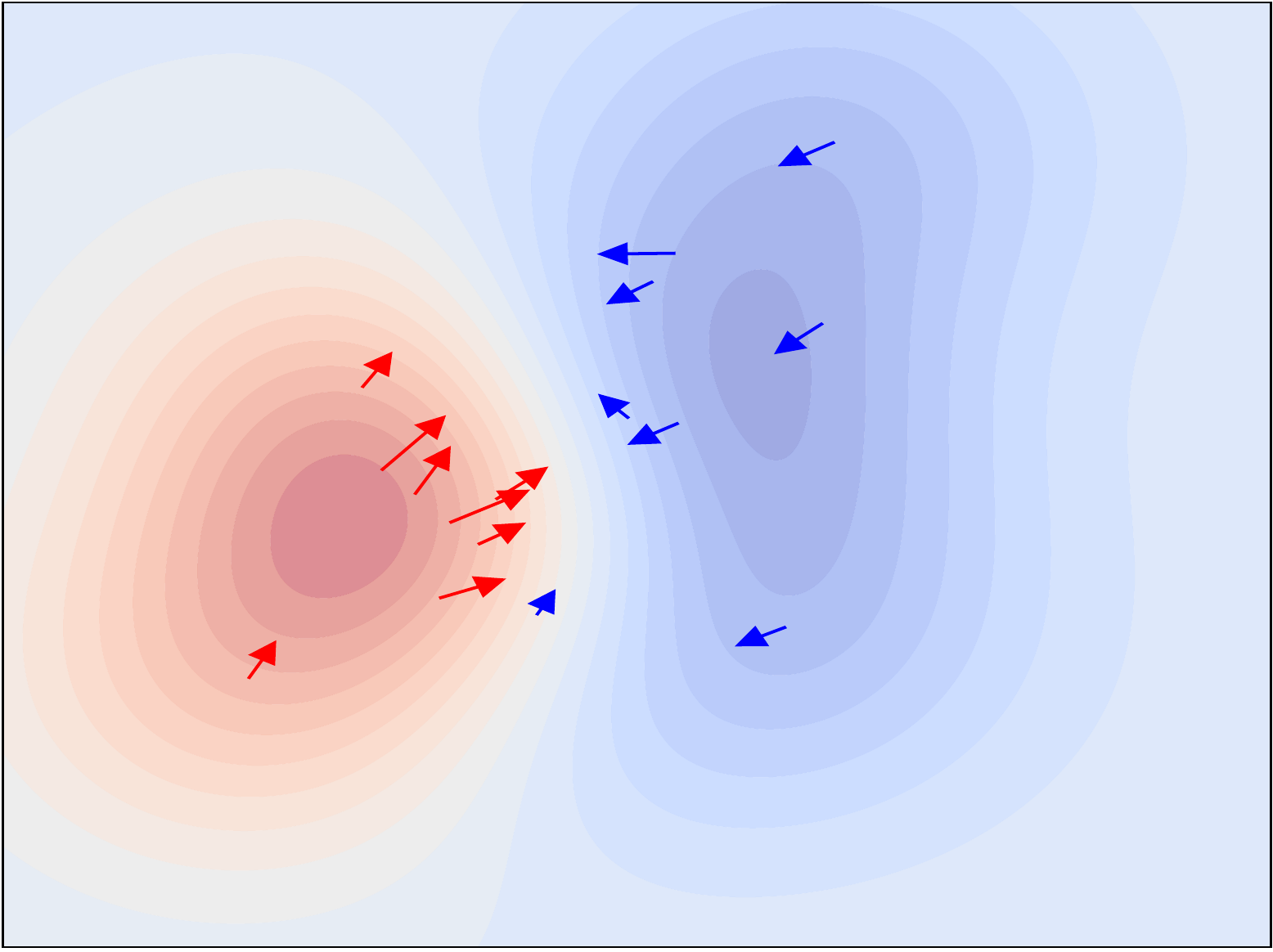}\label{fig:discriminative_CDM}}
   \caption{(a) While it is hard to assume a total order for models from different model classes under the oracle metric $E(p)$, it is plausible to assume partial orders for models from the same model class.  (b) Generative CDM uses the degraded distribution $p_n$ to synthesize fake samples for training a discriminator as the metric. The warm/cold region indicates the decision boundary of the resulting trainable metric induced by fake samples from $p_n$. (c) Discriminative CDM directly determines the decision boundary by pooling the values of the step-wise contrastive momentum.}  
\vspace{-15pt} 
\end{figure}

With the notion of oracle $E(p)$, we can perform: 
\begin{itemize}
    \item \emph{Discriminative} inference: \begin{itemize}
        \item[a)] \textbf{Distribution-level evaluation} to evaluate any existing models by ranking them according to $E(p)$ 
        \item[b)] \textbf{Sample-level evaluation} to use $\frac{\partial{E(p)}}{\partial p(\mathbf{s})}$ to reflect the \emph{quality} of $\mathbf{s}$. Because given the evaluated sequence $\mathbf{s}$, $\frac{\partial{E(p)}}{\partial p(\mathbf{s})}$ represents whether and how much altering the model $p$ towards higher $p(\mathbf{s})$ would improve $E(p)$. 
    \end{itemize} 
    \item \emph{Generative} inference: to improve or degenerate the generation quality by altering $p$ towards better or worse of $E(p)$. The altered distribution produces more obfuscating fake examples, which can then be used to train discriminator-based sample-level evaluation metrics.
\end{itemize}
In the following, we will explain discriminative and generative inference of CDM for automatically evaluation of open-domain generation in more details. 
 


 


\begin{figure*}[ht]   
  \centering  
    \subfigure[Generative CDM]{\includegraphics[height=0.45\columnwidth]{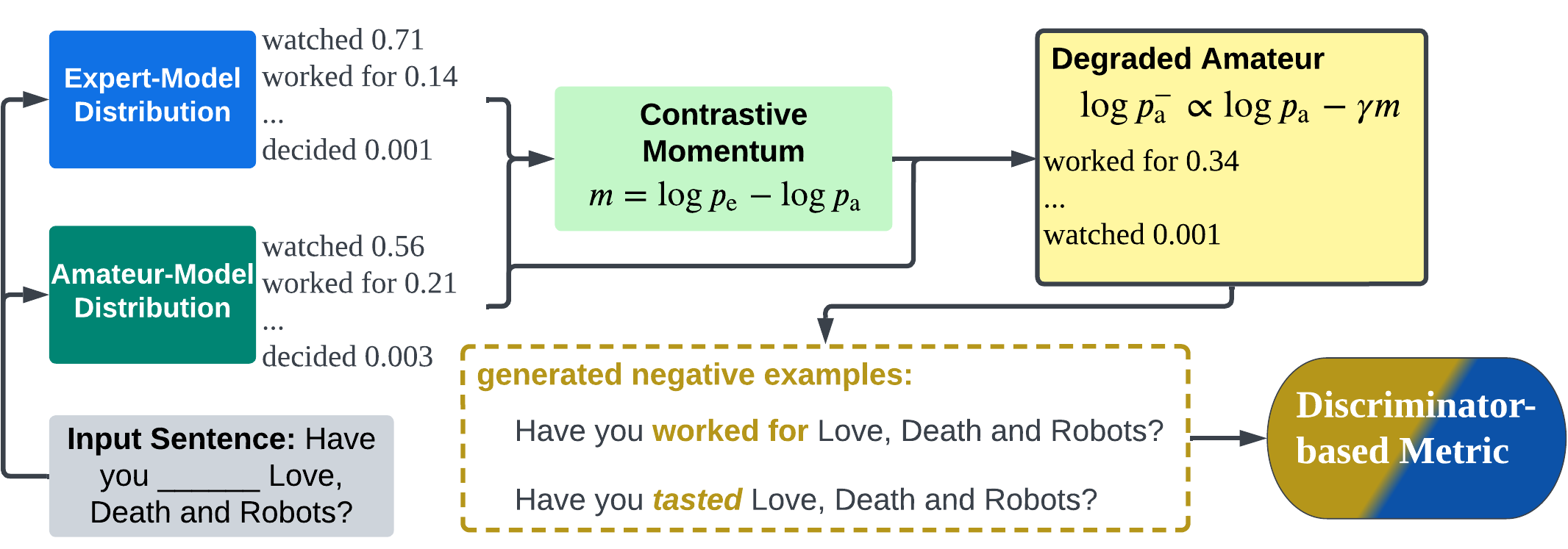} \label{fig:meta-distribution-g}} 
    ~\subfigure[Discriminative CDM]{\includegraphics[height=0.45\columnwidth]{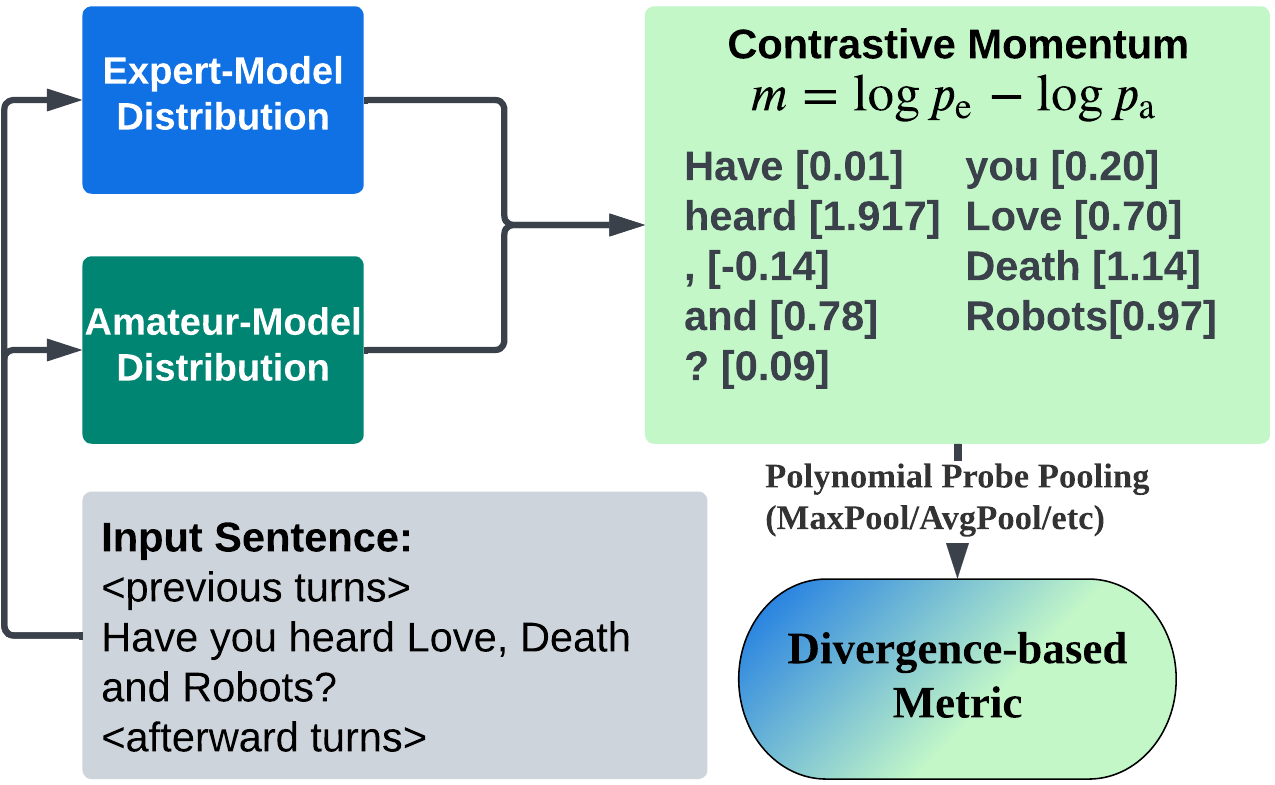}\label{fig:meta-distribution-d}}
  
  \caption{A more detailed illustration of the two Contrastive Distribution Methods (CDM). 
  (a) \emph{Generative CDM} constructs fake negative samples from positive ones for training a discriminator-based metric. (b) \emph{Discriminative CDM} directly evaluate the distribution/sequence by contrasting and aggregating the step-wise likelihood scores. }
  \vspace{-1.5em}
\end{figure*}

\subsection{The Partial Order Assumption} 
\sidistoryline{explain why implicit modeling is needed}

While it is nontrivial to come up with analytical forms for $E(p)$, we can make some assumptions to obtain partial orders from $E(p)$. 
Consider a \emph{series} of models that share similar architectures and other pretraining/finetuning setups, but differ in model sizes (e.g. T5-small/base/large, etc.). It is usually safe to assume that the model with a larger number of parameters perform better than the smaller one under most aspects. 
More formally, we can assume a partial order (a linear order within one concerned model class) induced by the oracle metric $E(p)$ as illustrated in Equation~\ref{eq:partial-order} and Figure~\ref{fig:secant}: 
\begin{equation}
\small
    E(p_{small}) < E(p_{base}) < E(p_{large}) \label{eq:partial-order}
\end{equation}

\paragraph{Limitation} Note that, while the partial order assumption is usually true for most existing model families in empirical practices, we are \emph{open} to the possibility that it \emph{might not hold} in some cases. As a result, the effectiveness of the proposed approach is inherently limited to cases where the partial order assumption holds.


\subsection{First Order Approximation of $E(p)$} 
Since we do not assume the knowledge about the analytical form of $E(p)$, it is intractable to compute $\frac {\partial{E(p)}}{\partial p(\mathbf{s})}$. 
However, following similar approach as in~\cite{li2022contrastive}, 
we can approximate $E(p)$ using a secant hyperplane between two distributions in the range of $E(p)$, i.e., the \emph{amateur} distribution $p_a$ and the \emph{expert} distribution $p_e$. 
In other words, we approximate $E(p)$ use the following analytic form: 
\begin{equation}
\small
E(p) = \sum _\mathbf{s} \Bigl( \log p_e(\mathbf{s}) - \log p_a(\mathbf{s}) \Bigr) p(\mathbf{s}),    \label{eq:e_grad} 
\end{equation}

It's trivial to prove that this approximation ensures $E(p_e) > E(p_a)$. We can further define the \textit{contrastive momentum}
$m(\mathbf{s}) = \log p_e(\mathbf{s}) - \log p_a(\mathbf{s})$. 
Different choices of $p_a$ and $p_e$ result in different \textit{contrastive momentum} and thus distinct quality of the first-order approximations for $E(p)$, hence different performance of the evaluation metric. We investigate the general principle for choosing the expert and amateur distributions in the experiment section.  

\subsection{Contrastive Distribution Methods}

\subsubsection{Generative CDM} 
\sidistoryline{Describe in detail how things work.}
Generative CDM focuses on synthetic data generation using contrastive distributions. We follow prior works such as ADEM~\citep{lowe2017autoturing} and DEAM~\citep{ghazarian2022deam} to formulate reference-free evaluation metrics as prediction tasks. 
In order to evaluate generated texts, a discriminator can be trained on positive and negative examples to serve as the evaluation metric.\footnote{The discriminator does not necessarily need to provide binary decisions, it can also produce scores. But we use binary examples for simplicity.} 
However, while we can assume human-written texts are positive examples, negative examples are non-trivial to obtain. 
Randomly generating negative examples using a uniform distribution over all possible sequences of tokens is not efficient, as most negative examples generated this way would be too trivial. On the other hand, generating negative examples by masking out spams in positive examples and having pretrained large-language models to fill in the masks may not result in real low-quality texts, which would confuse the discriminator. 

To this end, generative CDM provides a controllable approach to \textit{reduce} the quality of pretrained language models to generate ``deceptive negative examples''. 
Specifically, it generates from a ``novice'' distribution $p_n$ that descends along the direction of $-\frac {\partial{E(p)}}{\partial \log p}$ from the amateur model $p_a$ --- a weaker distribution than the amateur model.
Applying the approximation in Equation~\ref{eq:e_grad}, 
%
we follow the reversed direction of the contrastive momentum $m = \log p_e - \log p_a$ to degenerate from the \emph{amateur} model $p_a$. Mathematically, we obtain a probability distribution $\log p_n \propto \log p_a - \gamma m$ that \textit{amplifies} the likelihood of ``machine artifacts'' in a controllable (by setting the hyper-parameter $\gamma$) scale. Sampling from $p_n$ allows us to obtain suitable negative examples. 

\paragraph{Implementation Details.} We hereby discuss how to generate targeted negative examples. 
We start from existing positive examples $\mathbf{s}$ and construct the negative samples by masking out certain part $\mathbf{s}^{M+}$ of the positive ones, then conduct conditional generation using the remaining part $\mathbf{s}\setminus\mathbf{s}^{M+}$ as the initial context.
As a result, the generated negative examples would be more disguising compared to sampling directly from $p_n$. 
%
To achieve this, we train a \emph{segment infilling} model. Given a positive example and the position at which a segment is removed (randomly or strategically), we model the conditional distribution that reconstructs the original segment. 
We train an expert and an amateur model with segment infilling capabilities, we can then compose the distribution for sampling in the following form:

\vspace{-2em}
\begin{align}
    \log p_{edit}(\mathbf{s}^{M}|\mathbf{s}\setminus\mathbf{s}^{M+}) \propto& \log p_a(\mathbf{s}^{M}|\mathbf{s}\setminus\mathbf{s}^{M+}) \nonumber \\
    -& \gamma m(\mathbf{s}^{M}|\mathbf{s}\setminus\mathbf{s}^{M+}), \\
    \text{where~~}  
    m(\mathbf{s}^{M}|\mathbf{s}\setminus\mathbf{s}^{M+}) =& \log p_e(\mathbf{s}^{M}|\mathbf{s}\setminus\mathbf{s}^{M+}) \nonumber\\
    -& \log p_a(\mathbf{s}^{M}|\mathbf{s}\setminus\mathbf{s}^{M+}) \nonumber
\vspace{-1em}
\end{align}
This enables us to flexibly generate targeted negative examples that are deceptive. Figure~\ref{fig:meta-distribution-g} and \ref{fig:generative_CDM} illustrates this process in the procedural and distributional views. 

The full process of generative CDM can be summarized as follows:
\begin{algorithm}[htbp]
\caption{Generative CDM}
\begin{algorithmic}[1]
    \STATE Train the amateur model $p_a$ to solve the segment insertion problem
    \STATE Train the expert model $p_e$ to solve the segment insertion problem
    \STATE Construct the contrastive momentum $m_{a\rightarrow e} = \log p_e - \log p_a$
    \STATE Construct the degraded distribution $\log p_a^- \propto \log p_a - \gamma m_{a\rightarrow e}$
    \STATE negativeSamples = \{\}
    \FOR{positiveSample $\mathbf{s}^+$ \textbf{in} positiveSamples}
        \STATE Remove a segment $\mathbf{e}^+ \subset \mathbf{s}^+$ from $\mathbf{s}^+$ to construct the context $\mathbf{c} = \mathbf{s}^+-\mathbf{e}^+$
        \STATE Regenerate a segment $\mathbf{e}^-$ in the same position using $p_a^-(\mathbf{e}|\mathbf{c})$
        \STATE Obtain the reconstructed negative sample $\mathbf{s}^- = \mathbf{c} \cup \mathbf{e}^-$
        \STATE Add $\mathbf{s}^-$ to negativeSamples
    \ENDFOR
    \STATE Train the metric model $D$ as a discriminator with \{negativeSamples, positiveSamples\}
    \STATE \textbf{return} metric $D$
\end{algorithmic}
\end{algorithm}

\subsubsection{Discriminative CDM} 

Although Generative CDM is a reasonably flexible and scalable framework, 
there are many variable factors in the generation process (e.g. how to choose which segment to remove, the degradation strength factor $\gamma$ etc.) that may affect the performance of the resulting data and thus the evaluation metrics. 
Therefore, we propose an alternative paradigm under the CDM framework to remove the generation subroutine completely. 
In generative CDM, after data generation, we train a discriminator to distinguish positive and negative examples as the evaluation metrics. 
Effectively, we are learning the boundary between positive and negative samples, because we usually do not have a tractable model for the positive or negative distribution. 
However, under the CDM framework, we do have a tractable model for the negative distribution, which is composed from the amateur model $p_a$ and the expert model $p_e$. 
In light of this, we can consider directly deploying $m$ as a divergence-based metric for evaluation. 

For each sequence, we collect the step-wise contrastive momentum $m(x_t|\mathbf{s}_{<t})=\log p_e(x|\mathbf{s}_{<t}) - \log p_a(x|\mathbf{s}_{<t})$ composed from the \emph{amateur} model and the \emph{expert} model. 
For a good data sample, both models' likelihood prediction will be relatively high while the expert model will assign significantly higher probability to the sample, thus $\sum_t m(x|\mathbf{s}_{<t})$ should be significantly larger than 0. 

We can directly sum up the step-wise contrastive momentum over the entire sequence (\emph{i.e.} sum-pooling) to be the metrics for generation quality evaluation. See Figure~\ref{fig:meta-distribution-d} and \ref{fig:discriminative_CDM}. 
However, the sum-pooled score would be numerically influenced by the sequence length. Moreover, sum-pooling overemphasizes the impact of extremely low probability steps because the discrepancy between the amateur model and expert model predictions in low-probability regions could be significantly amplified on the logarithmic scale. 
Therefore, in the experiments, we compare different strategies to \emph{pool} the sequence of step-wise contrastive momentum values into a sentence-level evaluation score. We call this paradigm of using CDM as \emph{Discriminative CDM}, as we directly operate on two models' contrastive momentum on data samples as their quality evaluation metrics. 

\section{Experiments}

To support our claim that CDM is an evaluation metric that works generally for open-domain text generation tasks, we hereby conduct experiments under two distinct scenarios: 1) using CDM to evaluate dialogue systems, representing the tasks with a diverse distribution of outputs; 
2) using CDM to evaluate the commonsense of lexically-constrained generation models, representing tasks with medium to low diversity in the output. 
\subsection{Dialogue Evaluation} 
\label{sec:dialogue_eval}
The first part of our experiment is primarily focused on dialogue evaluation. Given a set of annotated dialogues, each with human-annotated quality scores ranging from $0.0$ to $1.0$, our objective is to assign scores to each evaluated sequence that maximizes the correlation with human annotations. Additionally, for dialogue evaluation, we assume we are not permitted to perform any training on data within the same domain.
Our training/fine-tuning exercises are conducted on a subset of dialogues from both TopicalChat \citep{Gopalakrishnan2019topical} and PersonaChat datasets \citep{zhang2018personalizing}, following the setup in \citet{ghazarian2022deam}. We evaluate our methods on annotated dialogues from the FED \citep{mehri2020unsupervised} and DSTC9 \citep{gunasekara2020overview} datasets.

In our experiment results, we report spearman correlation of different approaches for dialogue evaluation. All reported correlation coefficients from our approaches have $p$-value (with Bonferroni correction) $<$ 0.01. 

\paragraph{Dataset and Experiment Setup}
We adopt most experimental settings from DEAM \citep{ghazarian2022deam} to verify the effectiveness of our method. The statistics of the involved datasets in our experiments are shown as follows:
 
\begin{table}[h]
  \centering
  \small
  \vspace{-1.5em}
\caption{Data usage in the dialogue evaluation experiment.}
  \setlength{\tabcolsep}{0.5mm}
  \begin{tabular}{lcc}
    \toprule
    \textbf{Dataset} & \textbf{size} & \textbf{Avg. len}  \\
   \midrule 
   TopicalChat \citep{Gopalakrishnan2019topical}   & 17567 & \dual{377} \\
   + PersonaChat \citep{zhang2018personalizing} & +2078 & \\
   \midrule
   FED (test) \citep{mehri2020unsupervised} & 125 & 168 \\
   DSTC9 (test) \citep{gunasekara2020overview} & 2200 & 318 \\
  \bottomrule
  \end{tabular}
  \label{tab:data_usage}
  \vspace{-1.5em}
\end{table}

\begin{table}[h]
  \centering
  \caption{Main results of CDMs for dialogue evaluation in comparison to a few baselines. For G-Eval, to keep the comparison fair, we report results by instructing the largest models involved in CDMs. We include the original G-Eval metric result in our appendix. For UniEval, it is not capable of performing conversation-level evaluation, we report the average-pooled results of all utterance-level scores. Spearman correlation with human is reported. We highlight the best-performing results with \textbf{bolded} numbers and second-best with \ul{underlined} numbers. The models are selected using the validation set of Topical-Personal chat. 
  }
  \small 
  \setlength{\tabcolsep}{0.5mm}
  \resizebox{\columnwidth}{!}{
  \begin{tabular}{lcccc}
    \toprule
    \toprule
    \dual{Model} & \multicolumn{2}{c}{FED} & \multicolumn{2}{c}{DSTC9}\\ 
    & Coherence & Overall  & Coherence & Overall  \\ \toprule
    \citet{mesgar2019dialogue} & 0.10 & -0.01 & 0.02 & 0.05   \\ 
    \citet{vakulenko2018measuring} & 0.13 & 0.10 & 0.00 & 0.00   \\ 
    DynaEval \citep{zhang2021dynaeval} & -0.36 & -0.4 & -0.03 & -0.01   \\ 
    DEAM \citep{ghazarian2022deam} & 0.47 & \ul{0.55} & 0.19 & 0.20   \\
   
    \toprule
   G-Eval \citep{liu2023geval} & & & & \\
   \midrule
   - (w/ LLaMa2-7b-Vicuna) & \ul{0.57} & 0.54 & 0.15 & 0.14   \\
   - (w/ Flan-T5-11b) & 0.49 & 0.48 & 0.19 & 0.18   \\
   \toprule
   UniEval \citep{zhong2022unieval} & 0.33 & 0.35 & 0.14 & 0.13 \\\toprule 
    Generative CDM (Ours) &0.53 & \ul{0.55} & \ul{0.22} & \ul{0.24}   \\ 
   Discriminative CDM (Ours) &  \textbf{0.59} & \textbf{0.62} & \textbf{0.28} & \textbf{0.27}  \\
  \bottomrule
  \bottomrule
  \end{tabular}
  }
  
  \label{tab:dial}
  \vspace{-10pt}
\end{table}

\subsubsection{Model Specification}

For Generative CDM, there are multiple strategies to manipulate positive examples to generate negative examples. We study the following strategies: 
\begin{itemize}\itemsep0em 
    \item Segment/Utterance-Single: The manipulation of data is only applied once to a random \emph{segment no longer than 20 tokens} or \emph{a random utterance} in a real dialogue.
    \item Mixed-Single/Multi: The manipulation of data is applied to a random utterance or a random segment no longer than 20 tokens in a real dialogue, for once or a uniformly random value from 1 to 4 times.
    \item AMR-Multi: The location of data manipulation is guided by similar approach as in DEAM \cite{ghazarian2022deam}. 
\end{itemize}
Similarly, for discriminative CDM, there are many different aggregation strategies to pool step-wise contrastive momentums, and we study following:
\begin{itemize}\itemsep0em 
    \item Classifier-Pooled: We train a small linear classifier to convert the sequence of contrastive momentum scores as a trainable pooler using annotated training data (from the original dataset or as synthesized by DEAM \citep{ghazarian2022deam}). Intuitively, this is to align the aspect-agnostic \emph{contrastive momentum} with the concerned specific metric.
    \item Trivial pooling along the timestep axis (Avg-Pooled/Max-Pooled/Min-Pooled)
    
\end{itemize}


\subsubsection{Baselines}
Following the setup in \citet{ghazarian2022deam}, we compare against existing methods on negative sampling for trainable metrics, including \citet{mesgar2019dialogue}, \citet{vakulenko2018measuring}, DynaEval\citep{zhang2021dynaeval} and DEAM\citep{ghazarian2022deam}. We also report our comparison with some more advanced automatic evaluation metrics using pretrained large language models, including UniEval\citep{zhong2022unieval} and GEval\citep{liu2023geval}. For a fair comparison, we report GEval using instruction-tuned models no larger than the best \emph{expert} model involved in our CDM results.

  

\begin{table}[t]
  \centering
  \caption{Ablation studies focusing on manipulation strategies in generative CDM and pooling strategies in discriminative CDM with fixed armature model of T5-small and expert model of T5-large. Spearman correlation with humans is reported. All correlation coefficients from our approaches have $p$-value (with Bonferroni correction) $<$ 0.01.  
  }
  \small
  
  \setlength{\tabcolsep}{0.5mm}
  \begin{tabular}{lcccc}
    \toprule
    \toprule
    \dual{Model} & \multicolumn{2}{c}{FED} & \multicolumn{2}{c}{DSTC9}\\
    & Coherence & Overall  & Coherence & Overall  \\
   \toprule \multicolumn{5}{l}{Generative CDM (small-large)} \\
   \midrule 
    - Segment-Single & 0.12 & 0.07 & 0.11 & 0.10   \\
    - Utterance-Single & 0.29 & 0.36 & 0.05 & 0.08   \\
    - Mixed-Single  & 0.32 & 0.35 & 0.14 & 0.12   \\
    - Mixed-Multi  & 0.42 & 0.40 & 0.17 & 0.18   \\
    - AMR-Multi  & 0.49 & 0.53 & 0.20 & 0.22   \\

   \toprule \multicolumn{5}{l}{Discriminative CDM  (small-large)} \\
   \midrule
    - Avg-Pooled  & 0.31 & 0.32 & 0.12 & 0.13   \\
    - Min-Pooled  & 0.27 & 0.28 & 0.07 & 0.04   \\
    - Max-Pooled  & 0.46 & 0.43 & 0.16 & 0.15   \\
    - Classifier-Pooled  & 0.53 & 0.56 & 0.24 & 0.22  \\
  \bottomrule
  \bottomrule
  \end{tabular}
  \label{tab:abl_dialogue_manipulation}
  \vspace{-10pt}
\end{table}

\begin{table}[t]
  \centering
  \caption{Ablation studies focusing on varying amateur and expert model sizes in CDMs with the best manipulation strategy for generative CDM and the best pooling strategy for discriminative CDM. In addition to the T5 family, we report results of CDM using different sizes of LLaMa 2 as the amateur and expert models. Spearman correlation with humans is reported. All correlation coefficients have $p$-value (with Bonferroni correction) $<$ 0.01. 
  }
  \small
  
  \setlength{\tabcolsep}{0.5mm}
  \begin{tabular}{lcccc}
    \toprule
    \toprule
    \dual{Model} & \multicolumn{2}{c}{FED} & \multicolumn{2}{c}{DSTC9}\\
    & Coherence & Overall  & Coherence & Overall  \\
   \toprule Generative CDM &  &  &  &    \\
   \midrule 
    - T5 small-base & 0.48 & 0.51 & 0.19 & 0.20   \\
    - T5 small-large & 0.49 & 0.53 & 0.20 & 0.22   \\
    - T5 small-xl & 0.51 & 0.52 & 0.19 & 0.23   \\
    - T5 small-11b & 0.53 & 0.55 & 0.22 & 0.24 \\
    - T5 base-large & 0.29 & 0.31 & 0.08 & 0.09\\ 
    - T5 base-xl & 0.30 & 0.32 & 0.09 & 0.09 \\ 
    - T5 base-11b & 0.31 & 0.32 & 0.09 & 0.10 \\ 
    
   \toprule Discriminative CDM &  &  &  &    \\
   \midrule
    - T5 small-base & 0.42 & 0.44 & 0.13 & 0.10  \\
    - T5 small-large & 0.53 & 0.56 & 0.24 & 0.22  \\
    - T5 small-xl & 0.59 & 0.61 & 0.27 & 0.25  \\
    - T5 small-11b & 0.59 & 0.62 & 0.28 & 0.27  \\
    - T5 base-large & 0.39 & 0.40 & 0.09 & 0.11  \\
    - T5 base-xl & 0.47 & 0.46 & 0.12 & 0.13  \\
    - T5 base-11b & 0.52 & 0.51 & 0.15 & 0.16  \\
   
   \toprule \multicolumn{4}{l}{Finetuned LLaMa 2 7B-13B, No Infilling} \\
   \midrule
    - Generative CDM & 0.18 & 0.21 & 0.06 & 0.08 \\
    - Discriminative CDM & 0.20 & 0.22 & 0.13 & 0.15 \\
  \bottomrule
  \bottomrule
  \end{tabular}
  \label{tab:abl_dialogue_size}
  \vspace{-2.5em}
\end{table}

\begin{table}[t]
  \centering
  \caption{Comparisons between generative CDM and sampling directly from the amateur models. Spearman correlation with humans is reported. All correlation coefficients have $p$-value (with Bonferroni correction) $<$ 0.01. 
  }
  \small
  \setlength{\tabcolsep}{0.5mm}
  \begin{tabular}{lcccc}
    \toprule
    \dual{Model} & \multicolumn{2}{c}{FED} & \multicolumn{2}{c}{DSTC9}\\
    & Coherence & Overall  & Coherence & Overall  \\
   \midrule 
   \multicolumn{5}{l}{Generative CDM} \\ \midrule
    - T5 small-base & 0.48 & 0.51 & 0.19 & 0.20   \\
    - T5 small-large & 0.49 & 0.53 & 0.20 & 0.22   \\
    - T5 small-xl & 0.51 & 0.52 & 0.19 & 0.23   \\
    - T5 small-11b & 0.53 & 0.55 & 0.22 & 0.24 \\
   \midrule
   \multicolumn{5}{l}{Resampling from amateur models} \\ \midrule
    - T5-small & 0.31 & 0.28 & 0.09 & 0.08   \\
    - T5-base & 0.19 & 0.16 & 0.05 & 0.04   \\
    - T5-large & 0.09 & 0.05 & -0.01 & 0.02   \\
  \bottomrule
  \end{tabular}
  \label{tab:abl_dialogue_resample}
\end{table}

\begin{table}[h]
  \centering
  \caption{Comparisons between different manipulation capabilities in generative CDM. To achieve infilling with Pythia (an autoregressive model), we follow the fill-in-the-middle objective \citep{bavarian2022efficient}. Spearman correlation with humans is reported. All correlation coefficients have $p$-value (with Bonferroni correction) $<$ 0.01. 
  }
  \small
  \setlength{\tabcolsep}{0.5mm}
  \begin{tabular}{lcccc}
    \toprule
    \toprule
    \dual{Model} & \multicolumn{2}{c}{FED} & \multicolumn{2}{c}{DSTC9}\\
    & Coherence & Overall  & Coherence & Overall  \\
   \toprule \multicolumn{5}{l}{Manipulation using text infilling} \\
   \midrule 
    - T5 small-base & 0.48 & 0.51 & 0.19 & 0.20   \\
    - T5 small-large & 0.49 & 0.53 & 0.20 & 0.22   \\
    - T5 small-xl & 0.51 & 0.52 & 0.19 & 0.23   \\
    - T5 small-11b & 0.53 & 0.55 & 0.22 & 0.24 \\
   \midrule
    - Pythia 70M-160M & 0.28 & 0.31 & 0.04 & 0.06   \\
    - Pythia 70M-410M & 0.32 & 0.34 & 0.09 & 0.10   \\
    - Pythia 70M-1.0B & 0.35 & 0.37 & 0.10 & 0.11   \\
    - Pythia 70M-1.4B & 0.35 & 0.38 & 0.10 & 0.12 \\
    - Pythia 70M-6.9B & 0.44 & 0.41 & 0.18 & 0.22 \\
    
   \toprule \multicolumn{5}{l}{Manipulation using autoregressive generation} \\
   \midrule
    - Pythia 70M-160M & 0.26 & 0.27 & 0.02 & 0.04   \\
    - Pythia 70M-410M & 0.31 & 0.29 & 0.04 & 0.04  \\
    - Pythia 70M-1.0B & 0.36 & 0.37 & 0.06 & 0.05   \\
    - Pythia 70M-1.4B & 0.36 & 0.38 & 0.07 & 0.07   \\
    - Pythia 70M-6.9B & 0.41 & 0.43 & 0.09 & 0.08   \\
    
  \bottomrule
  \bottomrule
  \end{tabular}
  \label{tab:abl_dialogue_decoder_only}
  \vspace{-2em}
\end{table}
\subsubsection{Result Discussion and Detailed Ablation} 
Our approach aligns with methodology established by previous research and report the Spearman correlation to better evaluate CDM against these baselines. All hyperparameters for training the likelihood functions \emph{amateur} and \emph{expert} are determined with the best log-likelihood on the in-domain validation set from Topical-PersonaChat dataset.

\paragraph{Main Results} 
We report the main results with the two versions of CDM built from T5 models \citep{raffel2019t5,wei2021finetuned} checkpoints respectively in Table~\ref{tab:dial}. Discriminative CDM methods present less bias across datasets and offer more efficiency during training, as they eliminate the necessity for collecting negative samples and training an additional deep-NN-based classifier model. In general, Generative CDM performs the best among all trainable metrics using an explicit negative sampling process, while Discriminative CDM performs generally even better and achieve the state-of-the-art in all evaluated metrics. 

\paragraph{Ablation study: Negative sampling and pooling strategies} 
With the both versions of CDM being a composition of different components, we study how different components in our negative sampling strategy (for Generative CDM) and pooling (for Discriminative CDM) contribute to the performance of CDM metrics. See Table~\ref{tab:abl_dialogue_manipulation}. The performance of the metric obtained from Generative CDM is greatly impacted towards the negative sampling strategies. Manipulating the real samples in both utterance-level and segment-level for multiple times overall produces the best-quality negative samples. In addition, combining the idea in DEAM\citep{ghazarian2022deam}, we observe that further using AMR to perform the negative sample in a guided fashion is helpful, sufficiently making Generative CDM superior among all negative sampling-based metrics.

\paragraph{Ablation study: Impact of contrastive model sizes} We study how different model sizes in CDMs impact the performance of resulting metrics. See Table~\ref{tab:abl_dialogue_size}. Our findings indicate that larger performance gap between the amateur/expert models in general induces better performance. 

\paragraph{Ablation study: Generative CDM versus simply re-sampling from infilling models}  We also conduct experiments of directly synthesizing the negative samples by directly sampling from different sizes of amateur model only. We observe inverse scaling (i.e. sampling from better amateur model induces worse Generative CDM metric) in such attempts. This shows that contrasting the amateur model against the expert model is necessary for generating high-quality negative samples. See Table~\ref{tab:abl_dialogue_resample}.

\paragraph{Ablation Study: CDM with decoder-only models}  In Table~\ref{tab:abl_dialogue_decoder_only} we show results of CDM composed from decoder-only models. We report results with Pythia models in two different setups: 1) we finetine Pythia models in a regular paradigm 2) we finetune Pythia models with infilling capabilities using the fill-in-the-middle objective \citep{bavarian2022efficient}. In addition, we also report results with stronger decoder-only models like LLaMa 2 \citep{touvron2023llamatwo}. In general, infilling capabilities provide better flexibility in negative sampling as well as likelihood estimation with \emph{bi-directional} contexts, leading to better CDM metrics.

\subsection{Commmonsense Evaluation} 
In addition to the previous dialogue evaluation task, we consider a different setup where we use CDM to evaluate the commonsense of generated outputs in controllable generation problems. Generative commonsense reasoning has long been an interesting and challenging task, especially for models with smaller parameter numbers. We conduct this experiment not only to show that CDM can serve as a strong evaluation metric for commonsense evaluation, but also to demonstrate the generalizability of CDM. 

\subsubsection{Dataset Preparation}
CommonGen \citep{lin2019comgen} is a generative commonsense reasoning dataset that examines language models' capability of capturing commonsense and human logic. In CommonGen, the model is given a set of \emph{concept} keywords, and is expected to produce a descriptive sentence that: 1) contains all concept keywords (with necessary inflections for grammaticality); 2) compliant to commonsense. 

In the training split, CommonGen contains 32,650 unique concept sets, each with 1-3 annotated description. It also provides a validation set consisting of 992 unique concept sets, each with 4-5 annotated reference description. In test phase, there are 1496 compositionally varied concept sets that are intentionally made to be \emph{NOT} \emph{i.i.d.} to the training set. The golden annotations for these test input are not provided publicly.

\paragraph{Evaluating Commonsense Metrics with CommonGen-trinity} We reorganize and further annotate the test split of the dataset into a new dataset called \emph{CommonGen-trinity} to evaluate commonsense metrics. For each concept set in the test split, we use GPT-4 \citep{openai2023gpt4} to annotate 6 distinct descriptions containing every one of the keywords or its inflection. Furthermore, we prompt and control the large language model to produce samples with diverse degrees of commonsense: we generate 2 of the 6 annotations as fully commonsensical sentences; we control other 2 of the annotations to be of medium violation of commonsense, and 2 of the annotations to be completely non-commonsensical. See Table~\ref{tab:commongen_trinity}. We report the prompt (adapted from the ones in CommonGen-Lite\citep{lin2019comgen} for such fine-grained auto-annotation in our appendix.

During evaluation, we always annotate the highly commonsensical descriptions with a score of 1.0. We report the results under two setups for the annotation of other test descriptions:
\begin{itemize}
    \item \textbf{CommonGen-trinity-\emph{binarized}}: In this case, both \emph{mediocre} and \emph{non-commonsensical} descriptions are treated as non-commonsensical samples, and annotated with a score of 0.0.
    \item \textbf{CommonGen-trinity-\emph{raw}}: In this case we annotate \emph{mediocre} samples with a score of 0.5 and \emph{non-commmonsensical} samples with a score of 0.0. 
\end{itemize}

Similar to our setup in Section~\ref{sec:dialogue_eval}, for each baseline or model variant, we assign a score to each test description, and report the Spearman correlation to the golden scores of each sample.

\begin{table}[t]
  \centering
  \caption{Example of the proposed CommonGen-trinity dataset.
  }
  \small
  \setlength{\tabcolsep}{0.5mm}
  \begin{tabular}{ll}
    \toprule
    \toprule
    \textbf{Concept Set} & sidewalk dog walk leash \\
    \toprule
    \textcolor{MyDarkGreen}{\textbf{Commonsensical}} & A woman \ul{walks} her \ul{dog} on the \ul{sidewalk}, \\
    & holding tightly to the \ul{leash} \\
    \midrule
    \textcolor{MyDarkOrange}{\textbf{Mediocre}} & A \ul{dog} \ul{walks} its owner on a \ul{leash} along\\
    & the \ul{sidewalk}. \\
    \midrule
    \textcolor{MyDarkRed}{\textbf{Non-Commonsensical}} & A \ul{sidewalk} \ul{walks} a \ul{dog} with a \ul{leash} on \\
    & a dog. \\
  \bottomrule
  \bottomrule
  \end{tabular}
  \label{tab:commongen_trinity}
  \vspace{-1em}
\end{table}


\subsection{Results and Analysis}
We compare against G-Eval, the commonsense oracle built in BOOST \citep{tian2023boost} from COMET \citep{bosselut2019comet} and ACCENT \citep{ghazarian2023accent}. Following the setup in previous experiments, for a fair comparison, we report the G-Eval scores using instruction-tuned models no larger than the largest \emph{expert} model involved in CDMs.
\begin{table}[h]
  \centering
  \caption{Main results for evaluating the commonsense metrics using CommonGen-trinity dataset. Spearman correlation with the golden labels is reported. We highlight the best-performing results with \textbf{bolded} numbers and second-best with \ul{underlined} numbers. }
  \small 
  \setlength{\tabcolsep}{1mm}
  \begin{tabular}{lcc}
    \toprule
    \toprule
    Model & raw &  binarized \\ 
   
    \toprule
   G-Eval \citep{liu2023geval} & & \\
   \midrule
   - (w/ LLaMa2-7b-Vicuna) & 0.38 & 0.33 \\
   - (w/ Flan-T5-11b) & 0.61 & 0.53 \\
   \toprule
   \multicolumn{3}{l}{BOOST Commonsense Oracle \citep{tian2023boost}} \\
  \midrule
  - (w/ \emph{mean} aggregation) & 0.10 & 0.07 \\
  - (w/ \emph{total} aggregation) & 0.01 & -0.00 \\
   \toprule
   ACCENT \citep{ghazarian2023accent} & -0.08 & -0.07 \\
   \toprule 
    Generative CDM (Ours, small-11b) & \ul{0.63} & \ul{0.55}  \\ 
   Discriminative CDM (Ours, small-11b) & \textbf{0.73} & \textbf{0.61}  \\
  \bottomrule
  \bottomrule
  \end{tabular}
  
  \label{tab:commongen}
  \vspace{-2em}
\end{table}

\paragraph{Discussion} In general, CDM achieve the state-of-the-art among the evaluated metrics using a pretrained model with a maximal size of $11$b. G-Eval using the LLaMa2-7b model falls short for this task compared to results obtained from Flan-T5-11b, yielding that LLaMa2 may not be a better world model than Flan-T5. For BOOST Oracle, its detection of entity relation is limited to types in \textbf{UsedFor/AtLocation/CapableOf/PartOf}. Thus, while it is successful in guiding a generator, our results argue that it might not be the most competitive commonsense evaluation metric. For ACCENT, it is designed and trained for dialogue commonsense evaluation instead of single-sentence evaluation. To mitigate this, we report the results by composing a prompt containing problem descriptions as the virtual conversation history and evaluating the scene description as the next utterance. With these compromises, it is possible that the results for ACCENT may not reflect its true performance for the original use case. 

\section{Conclusion}
This paper presents the Contrastive Distribution Methods (CDM) as a general framework for evaluating open-domain text generation models. CDM is constructed around analyzing the correlation between model scales and the respective distribution prediction, and how it can be exploited to alter the performance of a certain model on-the-fly in inference. We demonstrate how CDM can be used for evaluation purposes in two general paradigms: Generative CDM, which manipulates existing positive samples to generate in-domain negative samples and subsequently trains a classifier, and Discriminative CDM, which employs the contrastive momentum as a direct metric for evaluation. Our experiments results in multi-turn dialogue evaluation and commonsense evaluation for controllable generation illustrate that CDM correlates better with human intuition than traditional metrics. In summary, the CDM method emerges as a promising and scalable approach for evaluating open-domain text generation systems, among others.

For future work, it is interesting to consider the contrastive momentum concerning more than two distributions as a reflection of an extended series of models across different scales. \looseness=-1

\section*{Limitation}
We hereby list a few potential limitations of the proposed method:
\begin{itemize} 
    \item While the method is proposed as a very general framework, training a metric with CDM is still a highly task-dependent practice. With our paper providing evidences for several principled practices (enlarging the discrepancies between models, etc), it could still need some efforts to apply CDM to a new task domain.
    \item It is theoretically infeasible for CDM to produce metrics for evaluating the inverse-scaling tasks (\emph{i.e.} the \textbf{bigger} the base model, the \textbf{worse} the performance), as it goes against the very basic assumption the CDM approaches rely on.
    \item Currently presented results of CDM are based on the linear, first order approximation of the oracle $E(p)$ using a secant hyperplane. This approach can be highly limited compared to a more accurate approximation. We leave this for future work.
\end{itemize}

\section*{Acknowledgement}
This research is supported by Meta Sponsor Research Award and Okawa Foundation Research Grant. We would also like to thank Sarik Ghazarian, Honghua Zhang, Meihua Dang and many other of UCLA/USC-PlusLab members for their kind writing suggestions and more. We also would like to express our deep appreciation to many of our anonymous reviewers for their insightful comments.
\newpage
\section*{Impact Statement}
The goal of this paper is to quantitatively study how the scaling-law of large language models correlate with human preferences without an explicit human-preference alignment process of any form. It might shed a light on a deeper understanding of the \emph{super-alignment} of language models, which means it could be possible that large language models achieve high agreement rate with human by simply using the feedback of language models with lesser capabilities.


\bibliography{MDM} 
\bibliographystyle{iclr2024_conference}
\appendix

\newpage
\section{Appendix}

\subsection{Prompt for G-Eval and Full Results with GPT-4 Model}

The prompts (for overall/coherence evaluation) we used for G-Eval are adapted from the original G-Eval repository. We hereby show case the one for dialogue overall score. Other prompts are mostly following the similar fashion.

\begin{mdframed}[backgroundcolor=gray!20]
You will be given a conversation between two agents.

Your task is to rate the dialogue on one metric.

Please make sure you read and understand these instructions carefully. Please keep this document open while reviewing, and refer to it as needed.

Evaluation Criteria:

Overall (1-10) - the overall quality of the whole dialogue. 

Evaluation Steps:

1. Read the dialogue carefully to get a general understanding of the overall quality of it.

2. Assign an overall score on a scale of 1 to 10, where 1 is the lowest and 10 is the highest based on the Evaluation Criteria.

Dialogue:

\{\textbf{The Dialogue}\}

Evaluation Form (scores ONLY):

- Overall:

......
\end{mdframed}
  \vspace{-2em}
\begin{table}[h]
  \centering
  \caption{All G-Eval results on the dialogue evaluation dataset. Spearman correlation with the golden labels is reported. We highlight the best-performing results with \textbf{bolded} numbers and second-best with \ul{underlined} numbers.
  }
  \small 
  \setlength{\tabcolsep}{0.5mm}
  \resizebox{\columnwidth}{!}{
  \begin{tabular}{lcccc}
    \toprule
    \toprule
    \dual{Model} & \multicolumn{2}{c}{FED} & \multicolumn{2}{c}{DSTC9}\\ 
    & Coherence & Overall  & Coherence & Overall  \\ \toprule
    \toprule
   G-Eval \citep{liu2023geval} & & & & \\
   \midrule
   - (w/ GPT-4, original) & \textbf{0.72} & \textbf{0.73} & \textbf{0.28} & \textbf{0.27}  \\
   - (w/ LLaMa2-7b-Vicuna) & \ul{0.57} & \ul{0.54} & 0.15 & 0.14   \\
   - (w/ Flan-T5-11b) & 0.49 & 0.48 & \ul{0.19} & \ul{0.18}   \\
  \bottomrule
  \bottomrule
  \end{tabular}
  }
  
  \label{tab:dial_geval}
  \vspace{-2em}
\end{table}

\begin{table}[h]
  \centering
  \caption{All G-Eval results for evaluating the commonsense metrics using CommonGen-trinity dataset. Spearman correlation with the golden labels is reported. We highlight the best-performing results with \textbf{bolded} numbers and second-best with \ul{underlined} numbers. }
  \small 
  \setlength{\tabcolsep}{1mm}
  \begin{tabular}{lcc}
    \toprule
    \toprule
    Model & raw &  binarized \\ 
   
    \toprule
   G-Eval \citep{liu2023geval} & & \\
   \midrule
   - (w/ GPT-4, Pseudo Upper Bound) & \textbf{0.81} & \textbf{0.67}  \\
   - (w/ LLaMa2-7b-Vicuna) & 0.38 & 0.33 \\
   - (w/ Flan-T5-11b) & \ul{0.61} & \ul{0.53} \\
  \bottomrule
  \bottomrule
  \end{tabular}
  
  \label{tab:commongen_geval}
  \vspace{-2em}
\end{table}
\newpage
\subsection{Prompt for CommonGen-trinity Synthesis}
\begin{mdframed}[backgroundcolor=gray!20]
\# Instruction \\
Given several concepts (i.e., nouns or verbs), write a short and simple sentence that contains *all* the required words. \\
With higher commonsense strength, the sentence should describe a more natural scene. \\
With lower commonsense strength, introduce more abnormal usages of the concepts or incorrect relations between them.\\
Make sure to generate as compact sentences as possible.\\
\# Examples \\
\#\# Example 1 \\
- Concepts: "dog, frisbee, catch, throw" \\
- Commonsense Strength: 5 out of 5\\
- Sentence: The dog catches the frisbee when the boy throws it into the air. \\
\\
\#\# Example 2 \\
- Concepts: "dog, frisbee, catch, throw" \\
- Commonsense Strength: 3 out of 5\\
- Sentence: A dog throws a frisbee at a dog as it tries to catch it.\\
\\
\#\# Example 3 \\
- Concepts: "dog, frisbee, catch, throw" \\
- Commonsense Strength: 1 out of 5\\
- Sentence: A dog throws a dog, while a frisbee trying to catch it.\\
\\
\# Your Task \\
- Concepts: "\{The concept set\}" \\
- Commonsense Strength: \{Commonsense Strength\} out of 5\\
- Sentence:
\end{mdframed}

\end{document}